\documentclass[11pt]{article}

\usepackage[preprint]{acl}
\usepackage{amsmath}
\usepackage{enumitem}
\usepackage{mathtools}
\usepackage{amssymb}
\usepackage{multicol, multirow}
\usepackage{booktabs}
\usepackage{float}
\usepackage{amsfonts}
\usepackage{times}
\usepackage{bbm}
\usepackage{latexsym}
\usepackage{footmisc}
\usepackage[T1]{fontenc}
\usepackage[T5]{fontenc}

\usepackage[utf8]{inputenc}
\usepackage{microtype}

\usepackage{inconsolata}

\usepackage{graphicx}
\usepackage{nicefrac}
\makeatletter
\newcommand{\blindfootnote}[1]{%
  \begingroup
  \renewcommand{\thefootnote}{}
  \footnotetext{#1}%
  \endgroup
}
\makeatother

\title{Echoes Across Vietnam's Highlands, Delta, and Coast: A Multilingual Corpus for Cham, Khmer, and Tay-Nung}

\author{
\textbf{Anh Trac Duc Dinh}$^{1,2,\dagger}$ \quad
\textbf{Khang Nhat Hoang Vo}$^{3,4,\dagger}$\thanks{Work done while visiting the National University of Singapore.} \quad
\textbf{Vinh Cong Doan}$^{2}$
\\
\textbf{Tai Tien Ta}$^{2}$ \quad
\textbf{Khoa Duc Anh Lam}$^{2}$
\\[0.6em]
$^{1}$Center for AI Research, VinUniversity, Hanoi, Vietnam
\\
$^{2}$Faculty of Computer Science and Engineering
\\
Ho Chi Minh City University of Technology (HCMUT), VNU-HCM, Vietnam
\\
$^{3}$Mohamed bin Zayed University of Artificial Intelligence, Abu Dhabi, UAE
\\
$^{4}$National University of Singapore, Singapore
\\[0.3em]
{\small
$^{*}$\textbf{Correspondence:}
\href{mailto:anh.dtd@vinuni.edu.vn}{anh.dtd@vinuni.edu.vn}
\quad
\href{mailto:Khang.Vo@mbzuai.ac.ae}{Khang.Vo@mbzuai.ac.ae}
}
}

\begin{document}
\maketitle
\blindfootnote{\textsuperscript{$\dagger$}These authors contributed equally to this work, and are project leaders. Names are ordered alphabetically.}

\begin{abstract}
Vietnam's ethnic minority languages are almost absent from the field of Natural Language Processing (NLP), and the challenge goes beyond data scarcity: Cham, Khmer, and Tay-Nung differ sharply in script, Vietnamese contact, and standardization, conditions under which standard multilingual adaptation can learn the wrong signals. We introduce CKTN, the first corpus and benchmark for these languages (44,367 documents, 24M subword tokens), spanning continued pretraining, category classification, and summary-document retrieval. We show that existing multilingual encoders severely fragment these languages, and that common adaptation metrics can mislead: models may lower language-modeling loss or excel at lexical-overlap retrieval while still failing at semantic generalization across documents. We address this with a script-aware adaptation recipe - vocabulary augmentation combined with calibrated replaced-token pretraining - that prevents the discriminator from exploiting trivial script mismatches. The result is an encoder with substantially less fragmentation and the strongest classification performance among evaluated models, exposing the limits of lexical-overlap retrieval as an evaluation signal.
\end{abstract}

\section{Introduction}

Low-resource NLP is often framed as a problem of missing data. For many minority languages, however, scarcity is only one part of the difficulty. Languages spoken under sustained contact with a dominant national language may diverge from their higher-resource genealogical relatives, develop unstable or mixed writing practices, and appear in digital text through scripts that are poorly represented by multilingual tokenizers. In such settings, adaptation can fail even when some text is available: models may learn local subword regularities, lexical overlap, or script cues rather than transferable language structure.

Vietnam's ethnic minority languages offer a compelling testbed for this problem. Vietnam recognizes 54 ethnic groups \citep{Van01082002,Tappe27052015}; beyond the Kinh majority, over 14 million people speak minority languages spanning the Austroasiatic, Tai-Kadai, Austronesian, and Sino-Tibetan families \citep{https://doi.org/10.1111/j.1749-818X.2007.00033.x,10.1093/molbev/msaa099}. We focus on three languages actively published in digital media yet almost absent from NLP: Cham, Khmer, and Tay-Nung. Together they form a natural contrast set: Cham (Austronesian, south-central coast) is traditionally written in a Brahmic-derived abugida but often appears in Latin script online; Khmer (Austroasiatic, Mekong Delta) uses a script without explicit word boundaries; and Tay-Nung (Tai-Kadai varieties of the northern highlands) uses Latin-based orthography but has limited standardized digital text. These languages thus span different regions, scripts, language families, and degrees of visible contact with Vietnamese.

This diversity is not merely sociolinguistic background; it creates concrete modeling failures. Existing multilingual encoders such as mBERT, XLM-R, and RemBERT \citep{pires-etal-2019-multilingual,conneau-etal-2020-unsupervised,DBLP:conf/iclr/ChungFTJR21} allocate little vocabulary capacity to these languages, fragmenting words into continuation pieces and weakening lexical representations. Moreover, script heterogeneity creates a specific hazard for ELECTRA-style continued pretraining \citep{clark2020electra}: a weak generator trained jointly across languages can produce replacements that are trivially detectable from script identity alone, such as substituting a Khmer-script token into a Latin-script Cham or Tay-Nung sequence. The discriminator can then solve replaced-token detection through shortcuts rather than semantic confusability.

We address these challenges by introducing CKTN, the first multilingual corpus and benchmark for Cham, Khmer, and Tay-Nung, and by proposing a script-aware adaptation recipe combining vocabulary augmentation with calibrated replaced-token pretraining. CKTN contains 44,367 documents and over 24M subword tokens, supporting continued pretraining, category classification, and summary-document retrieval. Our experiments show that tokenizer fragmentation strongly affects downstream transfer, that language-modeling loss and lexical-overlap retrieval can overestimate representation quality, and that calibrated script-aware pretraining substantially improves category classification.

Our contributions are:

\begin{itemize}[noitemsep,topsep=0pt]
    \item We introduce CKTN, the first corpus and benchmark for Cham, Khmer, and Tay-Nung (44,367 documents), supporting continued pretraining, 28-way category classification, and summary-document retrieval.
    \item We identify a failure mode of ELECTRA-style adaptation in script-heterogeneous settings: weak generators create script- or form-based corruptions that let the discriminator solve replaced-token detection through shortcuts.
    \item We propose a calibrated adaptation recipe-vocabulary augmentation \citep{gee-etal-2022-fast,minixhofer-etal-2022-wechsel}, linear generator-discriminator scheduling, and script-aware replacement filtering-that achieves the strongest classification performance, while our retrieval analysis explains why lexical-overlap tasks can overestimate representation quality.
\end{itemize}

\section{Related Work}

\subsection{Low-Resource Corpus for Vietnamese Ethnic Languages}
Corpus development for Vietnam's ethnic minority languages remains critically fragmented and scarce. While~\citet{le-etal-2004-spoken} and~\citet{trieu2020mtwiki} contributed foundational Vietnamese and Southeast Asian corpora, neither addressed indigenous minority languages. The few directly relevant efforts target only Bahnar:~\citet{Vo_Le_Phan_Nguyen_Pham_Tran_Nguyen_Vo_Quan_2024} applied NMT with transfer learning, and~\citet{nguyen-etal-2025-serving} introduced BARTBahnar with hybrid fine-tuning. Isolated attempts for a few other languages also exist, but only as small-scale dictionary-based or statistical systems published outside mainstream NLP venues, such as Vietnamese--Ede translation~\citep{le2015vietnamese} and Vietnamese--K'Ho translation~\citep{nguyen2023vietnamese}. This reflects a broader absence of coordinated NLP infrastructure, leaving dozens of Vietnam's minority languages without systematically collected, annotated, or task-ready corpora for NLP research.

\subsection{Tokenization as a Bottleneck for Low-Resource Language Modeling}

Multilingual models such as mBERT~\cite{pires-etal-2019-multilingual}, XLM-RoBERTa~\cite{conneau-etal-2020-unsupervised}, and RemBERT~\cite{DBLP:conf/iclr/ChungFTJR21} allocate vocabulary disproportionately toward high-resource languages, over-fragmenting low-resource languages and degrading performance. \citet{liang-etal-2023-xlm} quantify this directly: expanding XLM-R's vocabulary from 250K to one million tokens outperforms the original by 11.2\% on MasakhaNER and 5.8\% on Americas NLI. \citet{limisiewicz-etal-2023-tokenization} further confirm that language-specific token coverage, not model size, drives performance disparities on word-level tasks.

To address this, vocabulary adaptation methods transfer pretrained models to new languages without full retraining. FVT~\citep{gee-etal-2022-fast} replaces the tokenizer and reinitializes embeddings for efficient adaptation, while WECHSEL~\citep{minixhofer-etal-2022-wechsel} uses cross-lingual static embeddings to initialize target-language embeddings, enabling transfer of monolingual models to unseen languages. \citet{nag-etal-2023-entropy} show that entropy-based vocabulary augmentation yields significant gains for low-resource languages, collectively establishing vocabulary expansion as a necessary step in low-resource adaptation.

\subsection{Adapting Multilingual Pretrained Encoders to Low-Resource Languages}

Multilingual encoders such as mBERT and XLM-R, despite broad language coverage, often underperform on low-resource languages due to the \textit{curse of multilinguality}~\cite{chang-etal-2024-multilinguality}: as more languages share a fixed model capacity, per-language representation quality degrades. The most direct remedy is continued pre-training (CPT), where the model is further trained on target-language data. \citet{ebrahimi-kann-2021-adapt} show that CPT on limited New Testament data achieves substantial gains when adapting XLM-R to 1,600 languages, improving POS tagging by 17.69\% and NER by 6.29 F1 points, outperforming more complex methods in low-data conditions. \citet{downey-etal-2024-targeted} similarly demonstrate that targeted multilingual adaptation of XLM-R consistently outperforms mono- and multilingual baselines across 15 Uralic languages.

However, CPT risks overwriting multilingual knowledge through catastrophic forgetting. Modular approaches address this by isolating language-specific parameters. \citet{pfeiffer-etal-2022-lifting} introduce X-Mod, which adds language-specific modules to XLM-R, improving NER and question answering while allowing new languages to be added post-hoc without degrading existing ones. \citet{gurgurov-etal-2025-small} show that lightweight adapter methods, specifically Sequential and Invertible Bottleneck adapters, match or outperform full fine-tuning on NER and sentiment analysis using mBERT and XLM-R with far fewer trainable parameters. Together, these findings suggest that CPT is effective under extreme data scarcity, while modular adaptation is preferable when preserving cross-lingual transfer is a priority.

\subsection{Pretraining Objectives for Low-Resource Encoders: From MLM to ELECTRA}

Beyond vocabulary, the pretraining objective itself plays a critical role in
data efficiency, particularly when training signals are scarce. The standard
Masked Language Modeling objective learns from only 15\% of tokens per
step~\cite{devlin-etal-2019-bert}, leaving the majority of the input sequence unused. ELECTRA~\citep{clark2020electra}
addresses this inefficiency via Replaced Token Detection (RTD), a
discriminative objective that classifies every token as real or
generator-replaced, thereby utilizing the full sequence at each training
step. \citet{he-etal-2023-debertav3} demonstrate RTD's superior sample
efficiency: mDeBERTa Base achieves 79.8\% zero-shot accuracy on XNLI, a
3.6\% improvement over XLM-R. This efficiency advantage is especially
relevant in low-resource settings, where limited data makes every training
signal count. Indeed, \citet{cruz-etal-2021-exploiting} and
\citet{antoun-etal-2021-araelectra} successfully apply ELECTRA to Filipino
and Arabic respectively, both outperforming MLM-based baselines despite
constrained corpus sizes.

Crucially, these adaptations remain monolingual, single-script settings,
where sampling a replacement directly from the generator's raw output
distribution -- as in the original ELECTRA recipe -- is safe because the
generator is already well-calibrated. This assumption breaks down under
multilingual, script-heterogeneous, low-resource adaptation: a weak,
from-scratch generator can naively propose replacements that are wrong for
superficial reasons -- a Khmer-script piece substituted into a Latin
sequence, or a degenerate token standing in for a meaningful one -- letting
the discriminator learn shortcut cues (script, frequency) instead of
genuine semantic confusability, and eroding RTD's sample-efficiency
advantage. We take this as the core problem our method must solve, and
address it via a difficulty-calibrated replacement sampler that constrains candidates by script compatibility,
well-formedness, and embedding similarity -- the central mechanism enabling
ELECTRA-style pretraining to remain effective in this regime.

\section{Methodology}

\subsection{CKTN Corpus and Benchmark Design}

\begin{figure}[ht]
    \centering
    \includegraphics[width=\linewidth]{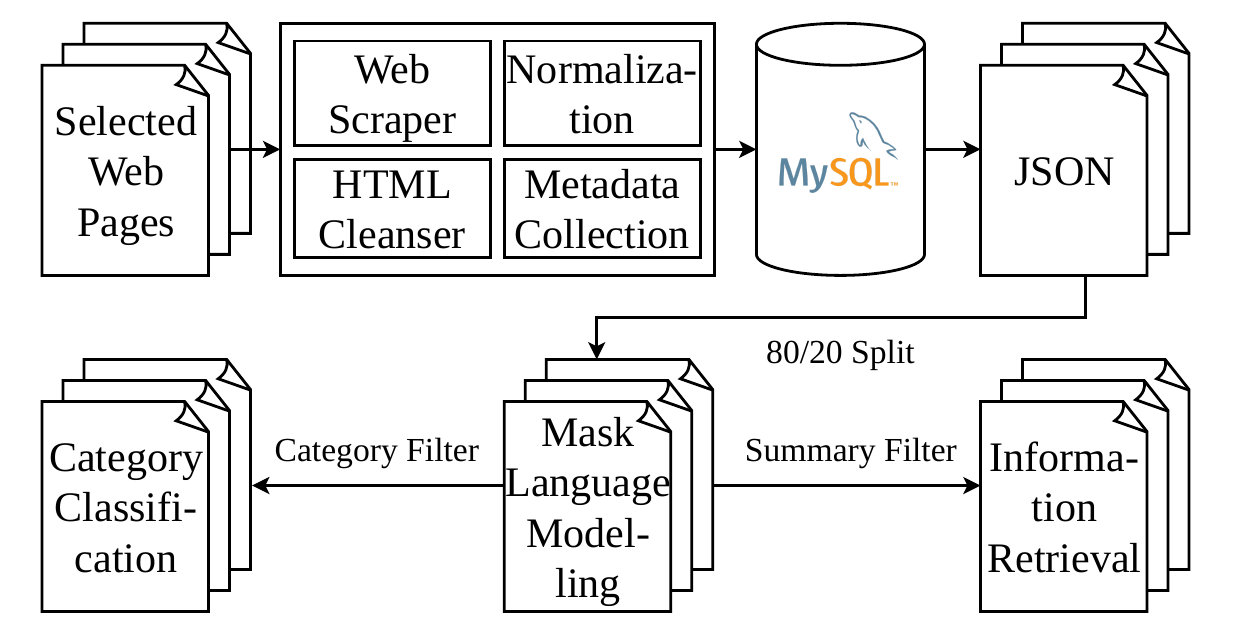}
    \caption{Data collection and preprocessing pipeline for minority language corpora (Cham, Khmer, Tay-Nung) from VOV and local news.}
    \label{fig:data_collection}
\end{figure}

CKTN is designed as both a resource for Vietnamese ethnic minority languages and a benchmark for low-resource transfer under script and contact variation. Unlike single-language low-resource collections, CKTN combines three languages differing in family, region, script, and Vietnamese contact: Cham (south-central coast), Khmer (Mekong Delta), and Tay-Nung (northern highlands), allowing us to test whether adaptation succeeds through lexical overlap, script compatibility, or genuine semantic transfer.

We collect data from continuously updated Vietnamese government and regional news outlets, including \textit{Voice of Vietnam},\footnote{\url{https://vov4.vov.vn/}} \textit{Ca Mau Newspaper},\footnote{\url{https://khmer.baocamau.vn/}} \textit{Can Tho Newspaper},\footnote{\url{https://baocantho.com.vn/khmer/}} \textit{An Giang Province Web Portal},\footnote{\url{https://angiang.gov.vn/khmer}} and \textit{Ethnic Minority and Mountainous Regions}.\footnote{\url{https://km-dantocmiennui.baotintuc.vn/}} The snapshot was collected around February 2026.

Figure~\ref{fig:data_collection} summarizes preprocessing: crawling, boilerplate removal, Unicode normalization, and metadata extraction (language, source, category, URL, title, summary, date). Khmer is additionally segmented with \texttt{khmer-nltk}, since its script lacks explicit word boundaries. Cleaned documents support three benchmark settings: continued pretraining, category classification, and summary-document retrieval.

The corpus contains 44,367 documents, 663,881 sentences, and over 24M BPE subword tokens (Khmer: 27,808; Cham: 11,481; Tay-Nung: 5,078 documents). Pretraining uses an 80/20 train-development split (seed 42). Classification uses the 9--10 most frequent categories per language (28 labels total). Retrieval uses documents with non-empty summaries, pairing each summary as a query with its article as the positive document. This design spans different levels of lexical dependence: pretraining evaluates token-level adaptation, retrieval measures recovery via shared surface forms, and classification tests generalization across lexically dissimilar articles. As our results show, models can excel at language modeling or lexical-overlap retrieval while still failing at cross-document semantic generalization.

\subsection{Vocabulary Augmentation}

The first obstacle in adapting multilingual encoders to CKTN is lexical fragmentation: existing multilingual tokenizers allocate little vocabulary capacity to Cham, Khmer, and Tay-Nung, splitting words into long sequences of continuation pieces. This weakens both downstream representations and ELECTRA-style pretraining, since fragmented lexical units make the generator more likely to propose malformed replacements, letting the discriminator rely on surface artifacts rather than semantic plausibility. We therefore expand the source vocabulary before continued pretraining.

Let $V_{\mathrm{src}}$ be the source encoder's original vocabulary and $C=\bigcup_{\ell} C_{\ell}$ the concatenated Cham, Khmer, and Tay-Nung pretraining corpora. We construct an extended vocabulary $V_{\mathrm{ext}}$ in three steps: candidate selection, tokenizer-score calibration, and embedding initialization.

\paragraph{Candidate selection.}
We train an auxiliary SentencePiece Unigram tokenizer \citep{kudo-richardson-2018-sentencepiece} on $C$ to obtain a candidate vocabulary $V_{\mathrm{aux}}$. A candidate piece $t \in V_{\mathrm{aux}}$ is retained if it is not already in $V_{\mathrm{src}}$, occurs at least $\tau$ times in the CKTN corpus, and passes a validity filter. The validity filter removes punctuation-contaminated pieces, control symbols, digit-dominant pieces, and forms that are inconsistent with the script profile of the language in which they occur. This prevents the extended vocabulary from being filled with boilerplate artifacts or cross-script noise.

\paragraph{Tokenizer-score calibration.}
SentencePiece Unigram tokenization performs Viterbi decoding over token log-scores. When new pieces are inserted, assigning them arbitrary scores can cause two failures: over-selection of long noisy pieces or under-selection of useful new pieces. We therefore initialize each new token score from its decomposition under the original tokenizer.

Let $I(t)$ be the sequence of source-vocabulary pieces produced by tokenizing the surface form $\tilde{t}$ with the original tokenizer, and let $\ell_{\mathrm{src}}(i)$ denote the original Unigram log-score of source piece $i$. We assign the new piece $t$ the calibrated score
\begin{equation}
\hat{\ell}(t)
=
\frac{1}{|I(t)|}
\sum_{i \in I(t)}
\ell_{\mathrm{src}}(i)
-
\lambda_{\mathrm{len}}(|\tilde{t}|-1)
\end{equation}
where $\lambda_{\mathrm{len}}$ is a length penalty. This places the new piece near the score of its original decomposition while discouraging overly long pieces. The resulting extended vocabulary is
\begin{equation}
V_{\mathrm{ext}} = V_{\mathrm{src}} \cup T,
\end{equation}
where $T$ is the set of retained CKTN pieces.

\paragraph{Embedding initialization.}
After extending the vocabulary, we expand the input embedding matrix while preserving the pretrained embedding space. Existing tokens keep their original embeddings. Following vocabulary-transfer and embedding-initialization approaches \citep{gee-etal-2022-fast,minixhofer-etal-2022-wechsel}, each new token $t$ is initialized from the embeddings of its source-token decomposition:
\begin{equation}
e_t =
\begin{cases}
\frac{1}{|I(t)|}\sum_{i \in I(t)} E^{(0)}_i,
& \text{if } |I(t)| > 0, \\[6pt]
\frac{1}{|J(t)|}\sum_{j \in J(t)} E^{(0)}_j,
& \text{if } |J(t)| > 0, \\[6pt]
\bar{e},
& \text{otherwise},
\end{cases}
\end{equation}
where $E^{(0)}$ is the original embedding matrix, $J(t)$ is a character-level decomposition of $\tilde{t}$ under the source tokenizer, and $\bar{e}$ is the mean source embedding. This initialization anchors new tokens in the existing multilingual space and avoids the instability of randomly initialized lexical entries.

The extended tokenizer is used for all subsequent continued pretraining and downstream evaluation. As shown in Section~\ref{sec:intrinsic}, this step substantially reduces fertility and continuation-token ratio across all three languages, giving the generator and discriminator cleaner lexical units for the calibrated pretraining stage.

\subsection{Script-Aware Calibrated Replacement Pretraining}
\label{sec:calibrated-rtd}

After vocabulary augmentation, we continue pretraining with an ELECTRA-style replaced token detection (RTD) objective, which supervises every non-padding token rather than only masked positions. Applying the standard recipe directly is unstable here due to two asymmetries: the discriminator is initialized from the vocabulary-augmented encoder while the generator is a much smaller from-scratch Transformer, and CKTN is script-heterogeneous (Cham/Tay-Nung mostly Latin script, Khmer in Khmer script). A weak generator can thus produce script-inconsistent replacements (e.g., a Khmer-script token in a Latin context), letting the discriminator solve RTD via script/formatting shortcuts instead of learning genuine language structure.

CKTN-ELECTRA addresses this with a smaller generator, a script-aware difficulty-calibrated replacement sampler, and a linear schedule for introducing the discriminator loss.

\paragraph{Generator objective.}
Given input $x=[x_1,\ldots,x_n]$ with masked view $x^m$ and original tokens $\hat{x}_i$, the generator $G_{\theta_G}$ is trained with standard MLM:
\begin{equation}
\mathcal{L}_{\mathrm{MLM}}(x;\theta_G) = \mathbb{E}\left[\sum_{i \in m} -\log p_G(\hat{x}_i \mid x^m)\right]
\end{equation}

\paragraph{Difficulty-calibrated replacement sampling.}
For each masked position $i$, we restrict candidates to the top-$k$ generator logits, then filter by four constraints: (1) not identical to the original token; (2) well-formed (no special/control tokens or punctuation-, digit-, symbol-dominant pieces unless the original is of the same class); (3) script-compatible with the local context (Khmer-script candidates excluded from Latin Cham/Tay-Nung contexts, and vice versa, barring script-agnostic symbols or named entities); (4) within an embedding-similarity band,
\begin{equation}
\rho_{\min} \leq \cos(e_{\hat{x}_i}, e_c) \leq \rho_{\max},
\end{equation}
which excludes candidates too similar (uninformative) or too dissimilar (trivially rejectable). Given the filtered set $V_i$, the replacement is drawn from a temperature-scaled softmax over generator logits restricted to $V_i$; if $V_i$ is empty, the token is left unchanged. RTD labels are defined relative to the original pre-masking token $\hat{x}_i$, so a generator sample equal to $\hat{x}_i$ is labelled unchanged.

\paragraph{Discriminator objective.}
Given corrupted sequence $x^c$ and label $y_t = \mathbbm{1}[x^c_t = \hat{x}_t]$, the discriminator $D_{\theta_D}$ is trained with binary cross-entropy over all non-padding positions:
\begin{align}
\mathcal{L}_{\mathrm{RTD}}(x;\theta_D)
&=
-\mathbb{E}\Bigg[
\sum_{t=1}^{n}
\Big(
y_t \log D_{\theta_D}(x^c,t)
\nonumber \\
&
+
(1-y_t)\log\big(1-D_{\theta_D}(x^c,t)\big)
\Big)
\Bigg]
\label{eq:rtd-loss}
\end{align}

\paragraph{Linear generator-discriminator scheduling.}
To avoid destabilizing the from-scratch generator, RTD is introduced gradually via $\mathcal{L} = \mathcal{L}_{\mathrm{MLM}} + \lambda(t)\mathcal{L}_{\mathrm{RTD}}$, with $\lambda(t)$ ramping linearly from 0 (warmup, generator-only) to $\lambda_{\max}$ (joint training) between $t_{\mathrm{warmup}}$ and $t_{\mathrm{ramp}}$, then held fixed.

\paragraph{Diagnostics and ablations.}
We track replacement rate, valid-candidate rate, RTD accuracy, discriminator confidence, and RTD entropy to confirm replacements are neither trivial nor excessively noisy. Ablations remove script-class constraints (sampling after only identity/well-formedness filtering) or the linear schedule (fixed $\lambda=\lambda_{\max}$ from step one). After pretraining, the generator is discarded and the discriminator is fine-tuned on downstream tasks.

\section{Experiments}

\subsection{Data Analysis}
We analyze token frequency and Zipfian distribution~\cite{zipf1935, zipf1949} on the pre-training MLM dataset, which serves as the foundation before splitting into downstream tasks; further data analysis is provided in Appendix~\ref{sec:data_analysis}.

Figure~\ref{fig:doc_length_dist_main} shows that Cham documents are substantially 
longer (peak $\approx$900 BPE tokens) than Khmer (peak $\approx$250) and 
Tay-Nung (bimodal, 50--500), reflecting differences in article style and 
orthographic density. Train and dev distributions align closely across all 
three languages, confirming that the splits are drawn from the same domain.

\begin{figure}
    \centering
    \includegraphics[width=\linewidth]{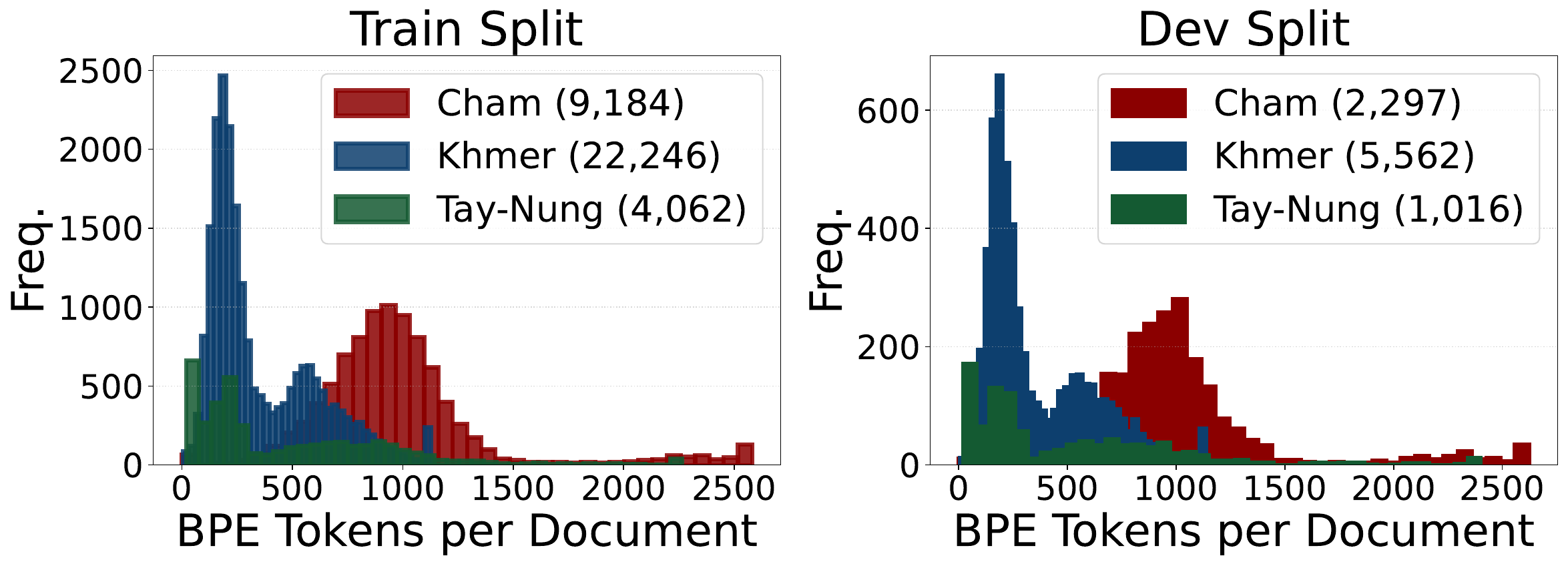}
    \caption{Distribution of BPE subword token counts per document across Cham, Khmer, and Tay-Nung MLM corpora.}
    \label{fig:doc_length_dist_main}
\end{figure}

\begin{figure}
    \centering
    \includegraphics[width=\linewidth]{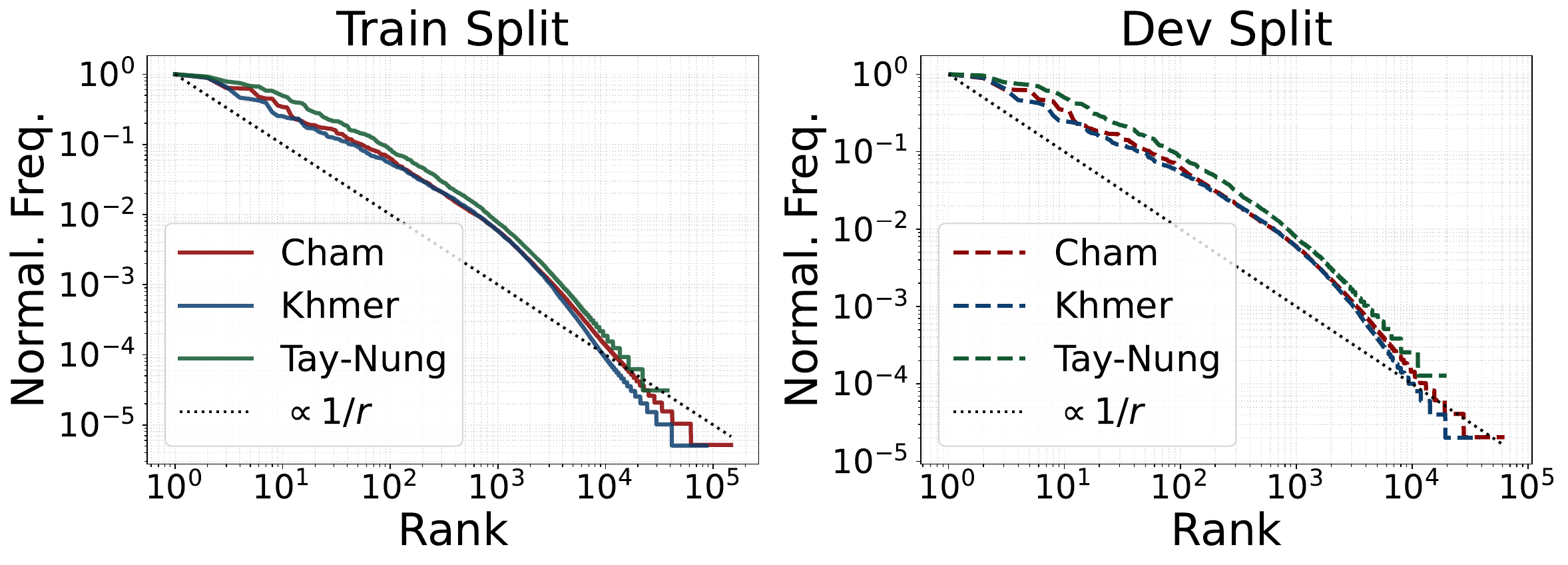}
    \caption{Word frequency distributions of the MLM corpora on a log-log scale, compared against the ideal Zipf law ($\propto 1/r$).}
    \label{fig:zipf}
\end{figure}

Figure~\ref{fig:zipf} confirms that all three corpora follow a natural Zipfian distribution, validating that the collected text is authentic and topically diverse rather than repetitive or synthetic. The close overlap between train and dev curves further indicates a consistent vocabulary distribution across splits, with no evidence of domain shift.

\subsection{Experimental Setup}
\label{sec:experimental-setup}

We evaluate three pretrained multilingual encoders: mBERT, XLM-R-base, and RemBERT, reporting performance before and after continued pretraining (CPT) on CKTN with the MLM objective for six epochs (15\% masking, maximum sequence length 512, learning rate $2 \times 10^{-5}$, warmup ratio 0.06, weight decay 0.01, and gradient clipping at 1.0). All experiments are conducted on a single 40GB GPU Tesla A100.

CKTN-ELECTRA initializes its discriminator from vocabulary-augmented RemBERT and pairs it with a generator at approximately one fourth of the discriminator's depth, trained from scratch jointly with the discriminator using a calibrated replacement procedure: top-$k=64$ candidate replacements, temperature $\tau=1.25$, and similarity band $[\rho_{\min}, \rho_{\max}] = [0.15, 0.95]$. The RTD weight $\lambda$ follows a linear schedule: $\lambda=0$ for epochs 1--2, increasing linearly to $\lambda_{\max}=50$ during epochs 2--3, and remaining fixed for epochs 4--6. After CPT, the generator is discarded and the discriminator encoder is fine-tuned for downstream tasks.

We evaluate models on three groups of metrics: (1) intrinsic tokenization quality, measured by fertility, token coverage, and continuation-token ratio; (2) category classification, evaluated using accuracy and Macro-F1 (the primary metric given the imbalanced label distribution), with all encoders fine-tuned for five epochs at maximum sequence length 512, selecting models on the development set and reporting held-out test performance; and (3) information retrieval, evaluated by encoding summaries as queries and articles as candidate documents, indexing embeddings with FAISS, and reporting MRR@10 and Recall@10. Since summaries and articles may share surface forms, retrieval results are interpreted together with classification rather than as a standalone measure of representation quality. For a detailed hyperparameters selection, please refer to Appendix \ref{app:reproducibility}.

\subsection{Intrinsic Evaluation}
\label{sec:intrinsic}
\begin{table*}[ht]
\centering
\renewcommand{\arraystretch}{0.85}
\setlength{\tabcolsep}{0.225cm}
\caption{Intrinsic evaluation results for multilingual model Tokenizer.}
\label{tab:tokenizer}
\begin{tabular}{c|l|c|c|c|c}
\toprule
\textbf{Language} & \textbf{Metric} & \textbf{CKTN-ELEC.} & \textbf{RemBERT} & \textbf{XLM-R} & \textbf{mBERT} \\
\midrule
\multirow{3}{*}{Cham}
  & Fertility $\downarrow$                        & \textbf{1.1824} & \underline{1.5862} & 1.6191 & 1.7576 \\
  & Token Coverage (\%) $\uparrow$                 & \textbf{7.3823}   & \underline{6.8113}   & 5.2734   & 6.3746   \\
  & Continuation Token Ratio (\%) $\downarrow$        & \textbf{4.8212}   & 36.7765  & \underline{22.6879}   & 35.1436   \\
\midrule
\multirow{3}{*}{Khmer}
  & Fertility $\downarrow$                        & \textbf{1.1721} & 1.6492 & 1.8365 & \underline{1.5037} \\
  & Token Coverage (\%) $\uparrow$                 & 11.7842           & 8.0324   & 6.9156   & \textbf{32.0145} \\
  & Continuation Token Ratio (\%) $\downarrow$        & \textbf{10.2238}  & 39.6847  & 34.5824  & \underline{26.6436}   \\
\midrule
\multirow{3}{*}{Tay-Nung}
  & Fertility $\downarrow$                        & \textbf{1.2414} & \underline{1.8021}  & 1.9186 & 1.9456 \\
  & Token Coverage (\%) $\uparrow$                 & \textbf{12.2723}  & \underline{10.7423}  & 9.8449   & 9.7154   \\
  & Continuation Token Ratio (\%) $\downarrow$        & \textbf{5.6142}   & 44.5425   & \underline{11.0311}  & 41.2632  \\
\bottomrule
\end{tabular}
\end{table*}

\begin{table*}[ht]
\centering
\renewcommand{\arraystretch}{0.6}
\setlength{\tabcolsep}{0.35cm}
\caption{MLM perplexity and loss before and after continued pretraining on the CKTN development set. CKTN-ELECTRA is excluded because it is trained with replaced token detection rather than MLM.}
\label{tab:mlm-cpt}
\begin{tabular}{c|l|c|c|c|c}
\toprule
\textbf{Condition} & \textbf{Metric} & \textbf{CKTN-ELECTRA} & \textbf{RemBERT} & \textbf{XLM-R} & \textbf{mBERT} \\
\midrule
\multirow{2}{*}{Before CPT} 
  & Perplexity $\downarrow$ & -- & 346,594.96 & \textbf{10.66} & \underline{27.45} \\
  & MLM loss $\downarrow$    & -- & 12.7559 & \textbf{2.3667} & \underline{3.3124} \\
\midrule
\multirow{2}{*}{After CPT}
  & Perplexity $\downarrow$ & -- & 4.68 & \underline{2.44} & \textbf{2.34} \\
  & MLM loss $\downarrow$  & -- & 1.5367 & \underline{0.8922} & \textbf{0.8505} \\
\bottomrule
\end{tabular}
\end{table*}

\begin{table*}[ht]
\centering
\renewcommand{\arraystretch}{0.85}
\setlength{\tabcolsep}{0.28cm}
\caption{Category Classification and Information Retrieval performance of various baselines before and after CPT on CKTN corpus. All baselines was vocabulary augmented before CPT and after CPT.}

\label{tab:classification_cktn}
\begin{tabular}{c|l|c|c|c|c}
\toprule
\textbf{Condition} & \textbf{Metric} & \textbf{CKTN-ELECTRA} & \textbf{RemBERT} & \textbf{XLM-RoBERTa} & \textbf{mBERT} \\
\midrule
\multicolumn{6}{c}{\textit{Category Classification}}\\
\midrule
\multirow{2}{*}{Before CPT} 
  & Accuracy $\uparrow$ & 0.0176 & 0.0176 & \textbf{0.7756} & \underline{0.7410} \\
  & Macro-F1 $\uparrow$ & 0.0007 & 0.0007 & \underline{0.3217} & \textbf{0.4615} \\
\midrule
\multirow{2}{*}{After CPT}
 & Accuracy $\uparrow$ & \textbf{0.9214}  &0.1454 &  \underline{0.8169} &0.8024  \\
 & Macro-F1 $\uparrow$ & \textbf{0.7103} & 0.0053  & \underline{0.5617} & 0.5054 \\
  
\midrule
\multicolumn{6}{c}{\textit{Information Retrieval}}\\
\midrule
\multirow{2}{*}{Before CPT} 
  & MRR@10 $\uparrow$ & \underline{0.7660} & \textbf{0.7727} & 0.7266 & 0.3628 \\
  & Recall@10 $\uparrow$ & \underline{0.9013}  & \textbf{0.9017} & 0.8712 & 0.4421 \\
\midrule
\multirow{2}{*}{After CPT}
  & MRR@10 $\uparrow$ & \underline{0.7718} & \textbf{0.7787} & 0.7361 & 0.3794 \\
  & Recall@10 $\uparrow$ & \underline{0.9039} & \textbf{0.9049} & 0.8789 & 0.4504 \\
\bottomrule
\end{tabular}
\end{table*}

Table~\ref{tab:tokenizer} reports fertility, token coverage, and continuation-token ratio across Cham, Khmer, and Tay-Nung. CKTN-ELECTRA achieves the lowest fertility for all three languages (1.18/1.17/1.24), a 25--32\% reduction over the strongest baseline, and cuts continuation-token ratio from as high as 44.5\% under RemBERT to below 11\%, showing that vocabulary augmentation gives the encoder cleaner lexical units. Token coverage also improves over RemBERT and XLM-R, except Khmer under mBERT, which reports higher coverage despite worse fertility and continuation-token behavior -- likely an artifact of character-level fallback rather than linguistically meaningful units. We therefore analyze coverage jointly with fertility and continuation-token ratio rather than in isolation.

Table~\ref{tab:mlm-cpt} reports MLM perplexity and loss before and after continued pretraining. Continued pretraining substantially reduces language-modeling loss for all baselines, showing that the CKTN corpus is sufficient to adapt existing multilingual encoders at the token-prediction level. However, the downstream results in Section~\ref{sec:extrinsic} show that low MLM loss is not by itself a reliable indicator of transferable document representations. In particular, an encoder may learn to predict frequent local subword fragments while still failing on tasks that require cross-document semantic generalization.

\subsection{Extrinsic Evaluation}
\label{sec:extrinsic}
Table~\ref{tab:classification_cktn} reports category classification and information retrieval results before and after continued pretraining, revealing sharply different behavior across the two tasks.

\paragraph{Category classification.}
CKTN-ELECTRA achieves the strongest performance after continued pretraining (0.9214 accuracy / 0.7103 Macro-F1), substantially outperforming XLM-R (0.8169 / 0.5617), mBERT (0.8024 / 0.5054), and RemBERT (0.1454 / 0.0053). Given the imbalanced label distribution, the Macro-F1 gain indicates transfer across minority classes rather than only dominant categories. RemBERT's collapse despite a large drop in MLM perplexity suggests a disconnect between intrinsic and extrinsic adaptation: without vocabulary augmentation, its tokenizer still fragments the target languages, so MLM improves local subword prediction without yielding topic-separable document representations - an effect potentially compounded by RemBERT's decoupled embeddings, which we leave for future mechanistic analysis.

\paragraph{Information retrieval.}
RemBERT remains strongest on MRR@10/Recall@10 both before and after CPT, while CKTN-ELECTRA improves only modestly - the opposite of the classification trend. We attribute this to retrieval's sensitivity to lexical overlap between summaries and their source articles, which can reward surface matching even when document-level semantic representations are weak. Classification and retrieval thus measure different forms of transfer: classification exposes cross-document semantic generalization, while retrieval reflects a mixture of semantic and lexical-overlap matching.

\paragraph{Why do XLM-R and mBERT outperform RemBERT under MLM?}
Since all three converge to similar post-CPT perplexity, the classification gap is likely architectural rather than optimisation-driven. Two non-exclusive hypotheses: (i) XLM-R and mBERT use tied, full-width embeddings (1024-d/768-d), so fine-tuning acts directly on the dimensions encoding token identity, unlike RemBERT's 256-d bottleneck; (ii) RemBERT's continuation-token ratio is especially high for Tay-Nung (44.5\% vs.\ 11.0\% for XLM-R), leaving more input as fragments with little standalone lexical signal. We cannot isolate which factor dominates.

\paragraph{CKTN-ELECTRA succeeds on the same backbone.}
Since CKTN-ELECTRA's discriminator shares RemBERT's architecture, its gains isolate the other two components. Vocabulary augmentation cuts fertility by 25--32\% and continuation-token ratio to under 11\%, giving the same 256-d embedding table cleaner, word-boundary-respecting entries. The ELECTRA objective further supervises every token rather than 15\%, likely valuable under our severe data scarcity. The calibrated sampler appears necessary: without script-compatibility and similarity constraints, replacements could be trivially detectable, letting the discriminator learn a shortcut instead of semantics. Together, these components jointly move the same RemBERT encoder from near-collapse to the best classification score.

\section{Ablation Study}

\begin{figure}[ht]
    \centering
    \includegraphics[width=\linewidth]{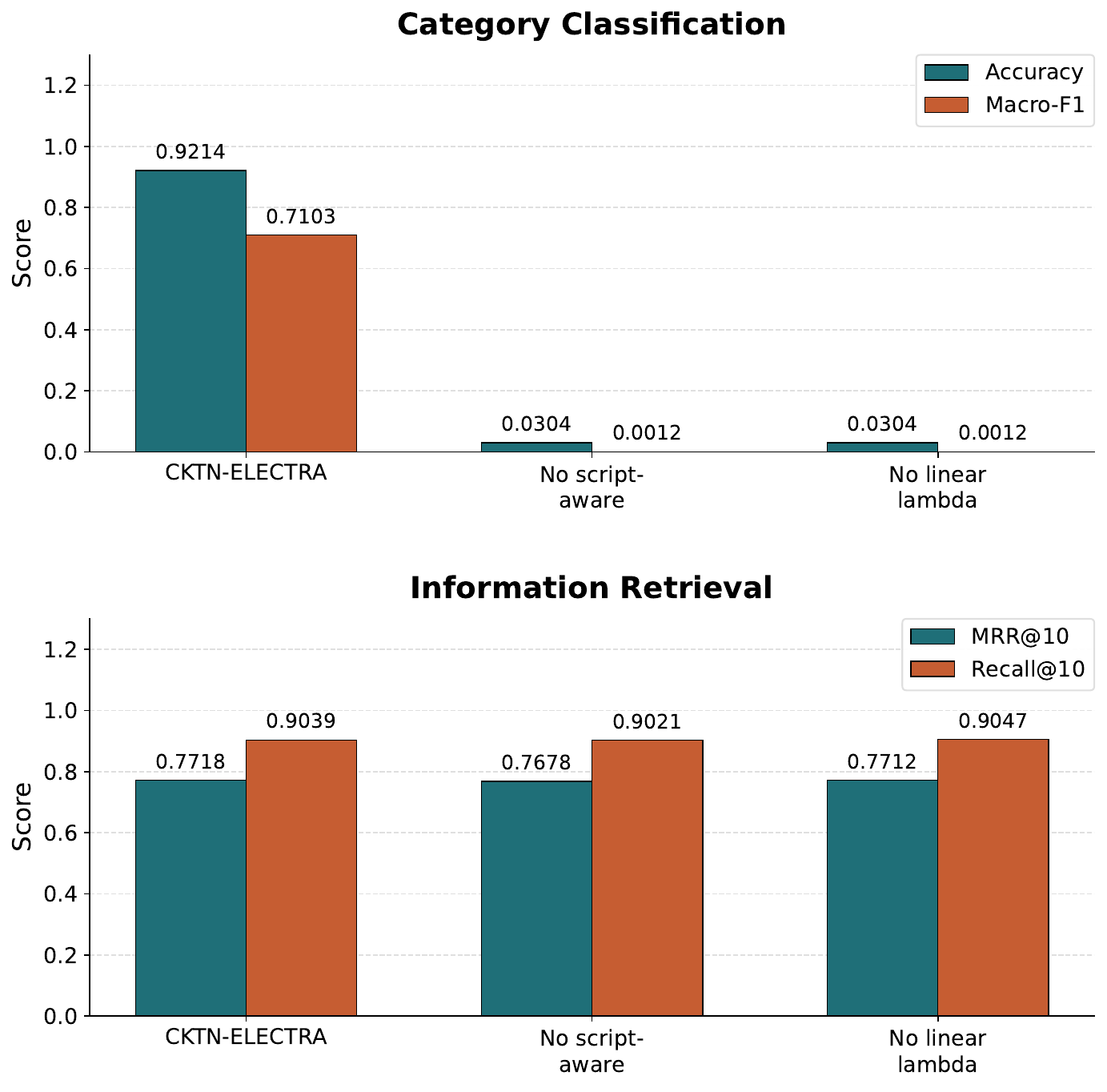}
    \caption{Ablation results for CKTN-ELECTRA.}
    \label{fig:ablation}
\end{figure}

\paragraph{Component importance.}
The ablation suggests script-aware filtering and linear scheduling are each important: removing either collapses classification to near-random levels (0.0304/0.0012 accuracy/Macro-F1 vs.\ 0.9214/0.7103 for the full model). One plausible explanation is that, without script-aware filtering, the from-scratch generator may substitute tokens from a mismatched script or of markedly different frequency, potentially letting the discriminator lean on such superficial cues rather than semantic confusability. Likewise, without linear scheduling, the full RTD loss is applied from the first update, pairing a strong pretrained discriminator with a generator that has not yet learned a plausible target-language distribution, which may destabilize training. We offer these as candidate explanations rather than confirmed mechanisms. What is clear is that retrieval remains largely unaffected in both ablations (MRR@10: 0.7678--0.7718; Recall@10: 0.9021--0.9047), indicating that classification, not lexical-overlap retrieval, is the metric sensitive to these components.

\section{Conclusion}
We introduced CKTN, the first multilingual corpus and benchmark for Cham, Khmer, and Tay-Nung, spanning continued pretraining, category classification, and summary-document retrieval. Because existing encoders severely fragment these languages and standard evaluation signals often overestimate representation quality, we developed CKTN-ELECTRA using a script-aware adaptation recipe that combines vocabulary augmentation with a difficulty-calibrated, script-constrained RTD objective. Ablations confirm that script-aware filtering and linear scheduling drive its superior classification performance, while our analysis exposes the limits of lexical-overlap retrieval as a standalone evaluation signal.

\section{Limitation}
Our work has several limitations. CKTN is sourced entirely from government and state-affiliated outlets, so it skews toward formal news register and away from informal or code-switched text, limiting generalization beyond institutional media. The category distribution is heavily imbalanced toward news, with labels inherited from source websites rather than independently verified. The corpus remains small overall (24M tokens, with Tay-Nung especially scarce), and the Khmer retrieval subset (250 pairs) is too limited for statistically reliable conclusions. Our explanations for RemBERT's collapse and for why script-aware filtering and linear scheduling matter are plausible hypotheses consistent with observed patterns, not confirmed via direct mechanistic probes. We evaluate only encoder-only backbones (mBERT, XLM-R, RemBERT), not decoder-based or instruction-tuned multilingual LLMs. Finally, since our retrieval task pairs summaries with their source articles, it inherently rewards lexical overlap, so results should not be read as a general-purpose semantic search benchmark.

\bibliography{custom}

\appendix

\section{Corpus Statistics}
\label{sec:appendix_A}

\begin{table*}[ht]
\centering
\renewcommand{\arraystretch}{0.8}
\setlength{\tabcolsep}{0.2cm}
\caption{Corpus statistics and task distribution}
\label{tab:data_statistics_updated}
\begin{tabular}{l@{\hspace{0.6cm}}|c|c|c|c}
\toprule
 & \textbf{Cham} & \textbf{Khmer} & \textbf{Tay-Nung} & \textbf{Total} \\
\midrule
\multicolumn{5}{l}{\textit{Source Distribution (Documents)}} \\
\midrule
 Voice of Vietnam (VOV)                               & 11,481 & 19,765 & 5,078 & 36,324 \\
 Ca Mau Newspaper\footnote{\url{https://khmer.baocamau.vn/}}                        &      0 &    537 &     0 &    537 \\
 Can Tho Newspaper\footnote{\url{https://baocantho.com.vn/khmer/}}                  &      0 &  6,609 &     0 &  6,609 \\
 An Giang Province Web Portal\footnote{\url{https://angiang.gov.vn/khmer}}          &      0 &    251 &     0 &    251 \\
 Ethnic Minority and Mountainous Regions\footnote{\url{https://km-dantocmiennui.baotintuc.vn/}} &      0 &    646 &     0 &    646 \\
\midrule
\multicolumn{5}{l}{\textit{Corpus Statistics}} \\
\midrule
 Documents           &     11,481 &     27,808 &      5,078 &     44,367 \\
 Sentences           &    301,317 &    294,077 &     68,487 &    663,881 \\
 Subword Tokens (BPE) & 11,361,144 & 10,174,988 &  2,495,431 & \textbf{24,031,563} \\
\midrule
\multicolumn{5}{l}{\textit{MLM Task Split (Documents)}} \\
\midrule
 Train               &      9,184 &     22,246 &      4,062 &     \textbf{35,492} \\
 Dev                 &      2,297 &      5,562 &      1,016 &      \textbf{8,875} \\
\midrule
\textit{Category Classification Task (Num Classes)} & 9 & 10 & 9 & 28 \\
\midrule
 Train               & 7,722 & 13,550 & 3,184 & 24,456 \\
 Dev                 & 1,408 & 2,351 & 556 & 4,315 \\
 Test                & 2,288 & 3,975 & 934 & 7,197 \\
\midrule
\multicolumn{5}{l}{\textit{Information Retrieval (Documents)}} \\
\midrule
 Summary Set              & 6,545 & 250 & 5,016 & 11,811 \\
 Document Set                & 6,545 & 250 & 5,016 & 11,811 \\
\bottomrule
\end{tabular}
\end{table*}

Table~\ref{tab:data_statistics_updated} presents the corpus statistics across three minority languages: Cham, Khmer, and Tay-Nung, totaling 44,367 documents, 663,881 sentences, and over 24 million BPE subword tokens. Khmer dominates the corpus with 27,808 documents sourced from VOV and four additional regional outlets, while Cham and Tay-Nung contribute 11,481 and 5,078 documents respectively.

The corpus is partitioned for two downstream tasks following an 80/20 split. For the Masked Language Modeling task, the full document set is divided into training and development subsets (35,492 and 8,875 documents respectively), preserving proportional language representation. For Category Classification, an additional held-out test set is introduced, yielding a three-way split across 28 classes in total (9--10 per language), with 24,456 training, 4,315 development, and 7,197 test documents. For Information Retrieval, matched summary--document pairs are used (11,811 pairs total); notably, the Khmer IR subset is substantially smaller (250 pairs) compared to Cham and Tay-Nung, reflecting limited availability of summary-level annotations in Khmer sources beyond VOV.

The data sources selected for this corpus share several defining characteristics that ensure both linguistic authenticity and long-term extensibility. \textit{Voice of Vietnam}, the primary source across all three languages, is a state-operated broadcaster with dedicated editorial teams comprising professional journalists and native-speaking contributors, guaranteeing consistent orthographic conventions and domain diversity. The supplementary Khmer sources such as \textit{Ca Mau Newspaper}, \textit{Can Tho Newspaper}, the \textit{An Giang Province Web Portal}, and the \textit{Ethnic Minority and Mountainous Regions portal} are likewise institutionally maintained, either by regional government bodies or by professional press organizations serving ethnic minority communities in the Mekong Delta. As active, continuously updated outlets, these platforms generate new content on a regular basis, meaning the corpus can be systematically expanded through periodic re-crawling without requiring new source identification or annotation pipelines. Furthermore, content produced by government portals and state-affiliated newspapers adheres to editorial standards that reduce noise from informal language use, code-switching, or user-generated irregularities commonly found in social med.

\section{Further Data Analysis}
\label{sec:data_analysis}


\subsection{Moving-Average Type Token Ratios (MATTR)}

\begin{table}[ht]
\centering
\footnotesize
\setlength{\tabcolsep}{3pt}
\begin{tabular}{lcccc}
\toprule
\textbf{Language} & \multicolumn{2}{c}{\textbf{Original RemBERT}} & \multicolumn{2}{c}{\textbf{Extended Tokenizer}} \\
\cmidrule(lr){2-3} \cmidrule(lr){4-5}
& \textbf{Train} & \textbf{Dev} & \textbf{Train} & \textbf{Dev} \\
\midrule
cham & 0.36727 & 0.36798 & 0.41432 & 0.41514 \\
khmer & 0.45615 & 0.45641 & 0.49069 & 0.49127 \\
tay\_nung & 0.31652 & 0.31488 & 0.39316 & 0.39078 \\
\bottomrule
\end{tabular}
\caption{Comparative MATTR results ($window=1000$) between the baseline and our extended tokenizer across all languages and splits.}
\label{tab:mattr-comparison-clean}
\end{table}
The comparative analysis of MATTR~\cite{Covington01052010} scores reveals a significant divergence between the Original RemBERT and our Extended Tokenizer across all evaluated languages. The results demonstrate a consistent and substantial increase in lexical diversity when employing the extended vocabulary; for instance, Tay-Nung exhibits the most pronounced improvement, with its MATTR rising from approximately $0.316$ to $0.393$. This upward trend indicates that the Original RemBERT tokenizer suffered from severe over-segmentation, breaking down words into repetitive and non-distinctive subword units, which artificially lowered the diversity metrics. In contrast, the higher MATTR values produced by the Extended Tokenizer suggest a more efficient encoding process that preserves lexical integrity, resulting in fewer repetitive functional subwords and a richer distribution of unique types per window. While the baseline scores were constrained by the specific sub-corpus characteristics and the limitations of the original multilingual vocabulary, our proposed extension successfully captures the specialized lexical nuances of Cham, Khmer, and Tay-Nung, providing a more robust and linguistically representative tokenization framework.
\subsection{Category Classification}
\subsubsection{Document Length}
\begin{figure}[!ht]
    \centering
    \includegraphics[width=\linewidth]{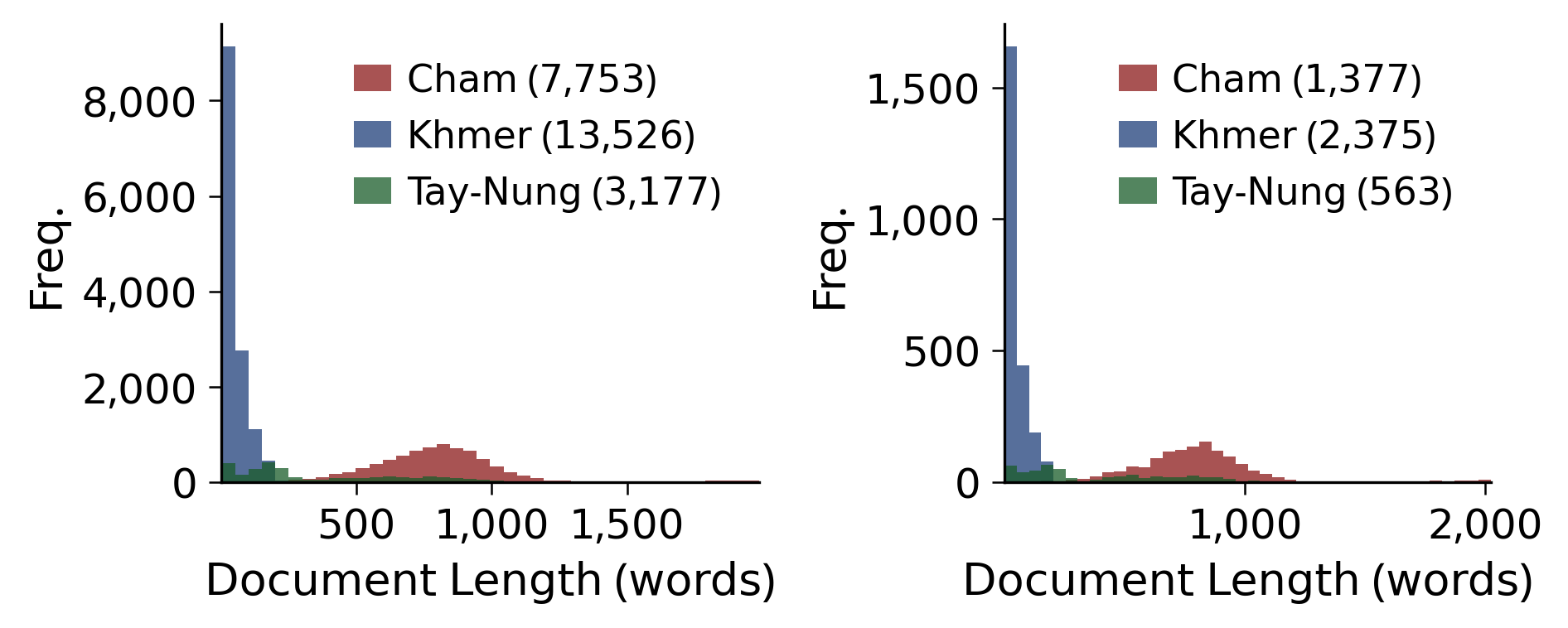}
    \caption{Document Length}
    \label{fig:doc_length_dist}
\end{figure}
The statistical distribution reveals that Khmer accounts for the largest portion of the dataset in terms of document counts, with most samples being relatively short. There is a high degree of consistency in the document length patterns between the Train and Dev splits, ensuring that the development set is statistically representative of the training data. Notably, while Khmer and Tay-Nung documents are primarily concentrated at shorter lengths, the Cham sub-corpus exhibits a distinct distribution, with most documents ranging from $500$ to $1000$ words. This variation suggests that the model must handle significantly different levels of contextual density across the three languages.
\subsubsection{Document sentences}
\begin{figure}[ht]
    \centering
    \includegraphics[width=\linewidth]{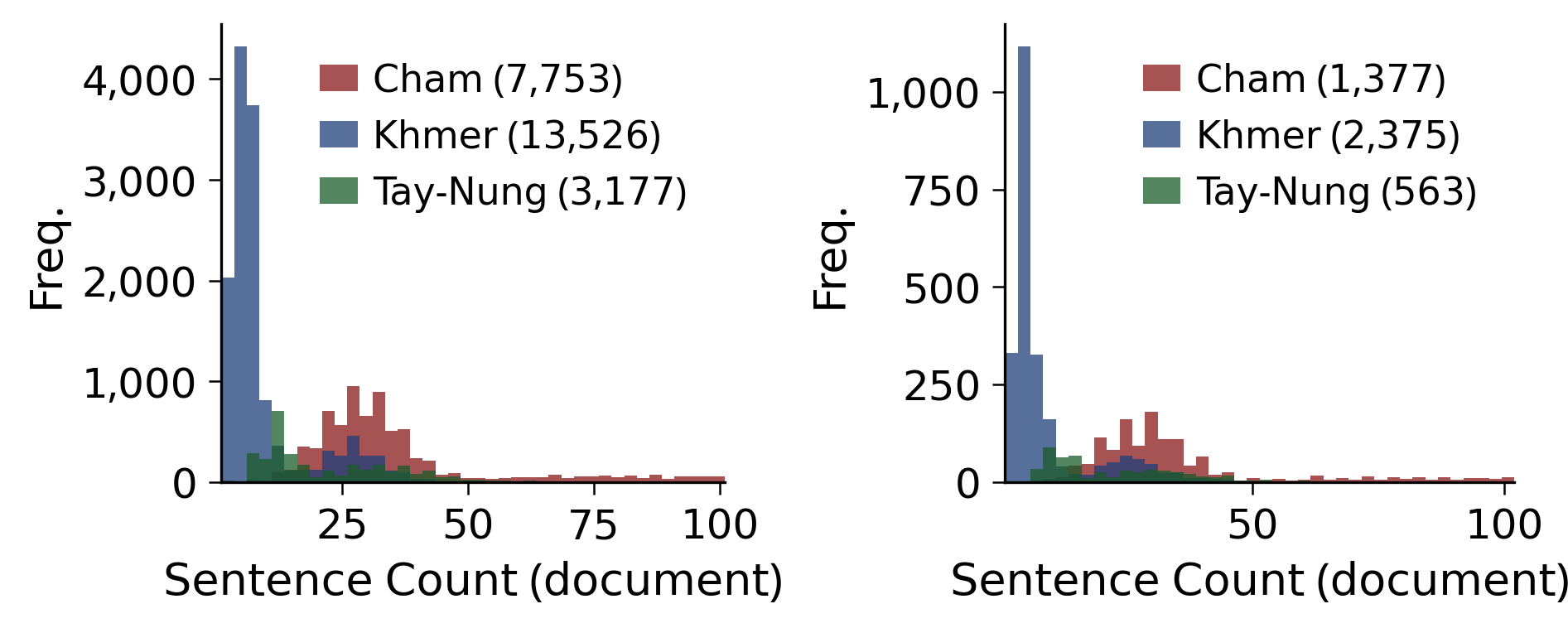}
    \caption{Document Sentences}
    \label{fig:doc_sentence}
\end{figure}

 The distribution of sentence counts per document shown in Fig.~\ref{fig:doc_sentence} further reinforces the structural differences between the three languages. Khmer documents are characterized by a high frequency of very short structures, with most containing fewer than $10$ sentences. In contrast, the Cham sub-corpus demonstrates a more balanced and distributed pattern, with a peak around $20 - 40$ sentences per document. The consistent overlap between the Train and Dev sets across all languages confirms that the data splits maintain a stable sentence-level complexity. These observations indicate that while Khmer provides a larger volume of data, the Cham documents offer more extended intra-document sentence sequences, which could be a significant factor for models capturing long-range dependencies.

\subsubsection{Category Distribution}
\begin{table}[ht]
\centering
\small

\label{tab:top5_categories}
\begin{tabular}{llr}
\toprule
\textbf{Language} & \textbf{Category} & \textbf{Count} \\
\midrule
\multirow{5}{*}{Cham} & News \& Events & 4,352 \\
 & Cham Community Issues & 2,663 \\
 & Ethnic Groups of Vietnam & 463 \\
 & Your Doctor & 451 \\
 & Law \& Life & 407 \\
\midrule
\multirow{5}{*}{Khmer} & News & 13,182 \\
 & Discovering Vietnam & 570 \\
 & People of Vietnam & 462 \\
 & Daily Reports & 461 \\
 & Vietnamese Economy & 417 \\
\midrule
\multirow{5}{*}{Tay-Nung} & Daily Radio Program & 1,248 \\
 & News \& Events & 951 \\
 & Tay-Nung Music & 397 \\
 & New Music & 233 \\
 & Tourism & 170 \\
\bottomrule
\end{tabular}
\caption{Top 5 categories per language (Train + Dev combined)}
\end{table}
The category distribution across all three languages reveals a strong dominance of news and current events content. For Cham, the two largest categories such as News \& Events (4,352) and Cham Community Issues (2,663), together account for the majority of the sub-corpus, reflecting the journalistic and community-oriented nature of the data source. A similar pattern is observed in Tay-Nung, where News \& Events (951) and Daily Radio Program (1,248) dominate, suggesting that the corpus is primarily drawn from broadcast and news media. For Khmer, the News category overwhelmingly leads with 13,182 documents, representing the vast majority of the Khmer sub-corpus. In contrast, culturally-oriented topics such as Tourism, Music, and New Music appear only marginally across all languages, indicating that modern lifestyle and entertainment content remains underrepresented in the dataset. This skewed distribution suggests that while the corpus is well-suited for news-domain tasks, it may not fully capture the breadth of informal or cultural language use in these minority communities.

\subsection{Vietnamese Loanword Penetration Analysis}

We measured the influence of Vietnamese on the three minority languages by tracking the hit rates of common Vietnamese administrative loanwords:
\begin{itemize}[noitemsep,topsep=0pt]
    \item \textbf{Cham (High Borrowing):} Shows the highest penetration of Vietnamese entities (e.g., \textit{``Việt Nam''} at 25.59\% and \textit{``y tế''} at 20.26\%). This suggests a hybrid linguistic profile where official and health-related terminology is heavily borrowed directly from Vietnamese.
    \item \textbf{Tay-Nung (Moderate Borrowing):} Despite sharing the Latin script, borrowing rates are moderate compared to Cham (e.g., \textit{``Việt Nam''} at 18.24\%, \textit{``Chính phủ''} at 6.06\%). This shows a degree of lexical overlap with Vietnamese,  due to shared administrative and political vocabulary in broadcast media.
    \item \textbf{Khmer (Zero Direct Borrowing):} The 0\% hit rate across all keywords confirms that the script barrier prevents any direct orthographic borrowing from Vietnamese, despite potential phonetic borrowing at the spoken level.
\end{itemize}

The selected entities include:
\begin{itemize}[noitemsep,topsep=0pt]
    \item \textbf{Political/Administrative:} \textit{Việt Nam} (Vietnam), \textit{Chính phủ} (Government), \textit{Thủ tướng} (Prime Minister), \textit{Bộ Giáo dục} (Ministry of Education).
    \item \textbf{Social/Functional:} \textit{văn hóa} (Culture), \textit{phát triển} (Development), \textit{y tế} (Health), \textit{quy định} (Regulation).
\end{itemize}

\begin{table}[H]
\centering
\small 
\begin{tabular}{lrr}
\toprule
\textbf{Entity} & \textbf{Tay-Nung (\%)} & \textbf{Cham (\%)} \\
\midrule
Việt Nam      & 915 (18.24) & 1,675 (25.59) \\
Chính phủ     & 304 (6.06)  & 70 (1.07)    \\
văn hóa       & 275 (5.48)  & 105 (1.60)   \\
phát triển    & 271 (5.40)  & 169 (2.58)   \\
y tế          & 221 (4.41)  & 1,326 (20.26) \\
Thủ tướng     & 136 (2.71)  & 853 (13.03)  \\
quy định      & 57 (1.14)   & 239 (3.65)   \\
Bộ Giáo dục   & 6 (0.12)    & 8 (0.12)     \\
\bottomrule
\end{tabular}
\caption{Distribution of Vietnamese administrative entities in Tay-Nung and Cham sub-corpora. }
\label{tab:entity-dist}
\end{table}

\begin{figure}[ht]
    \centering
    \includegraphics[width=\linewidth]{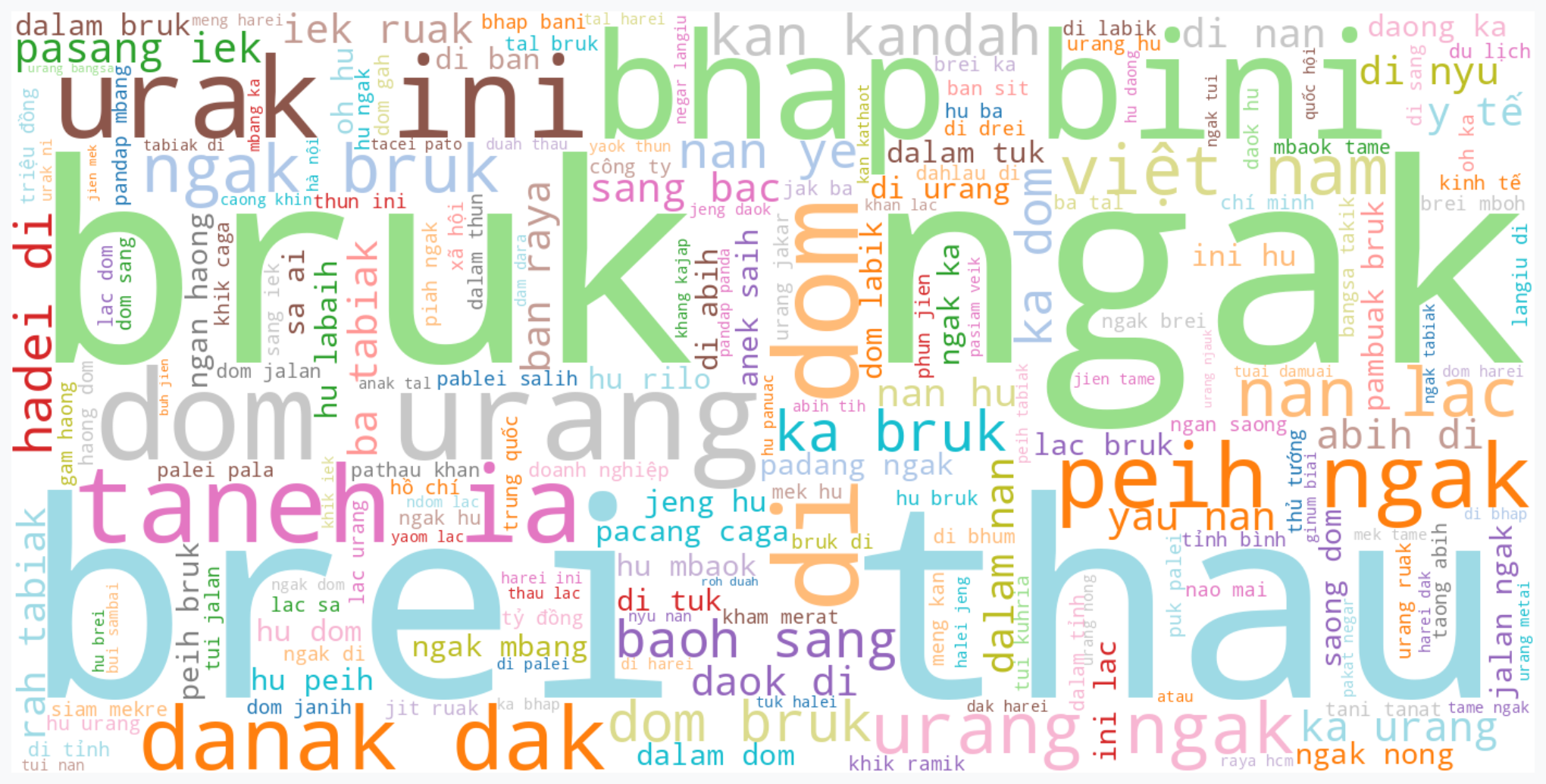}
    \caption{Word cloud of most frequent tokens in the Cham corpus, dominated by words such as \textit{bruk}, \textit{ngak}, \textit{thau}, and \textit{brei}.}
    \label{fig:wordcloud_cham}
\end{figure}
\begin{figure}[ht]
    \centering
    \includegraphics[width=\linewidth]{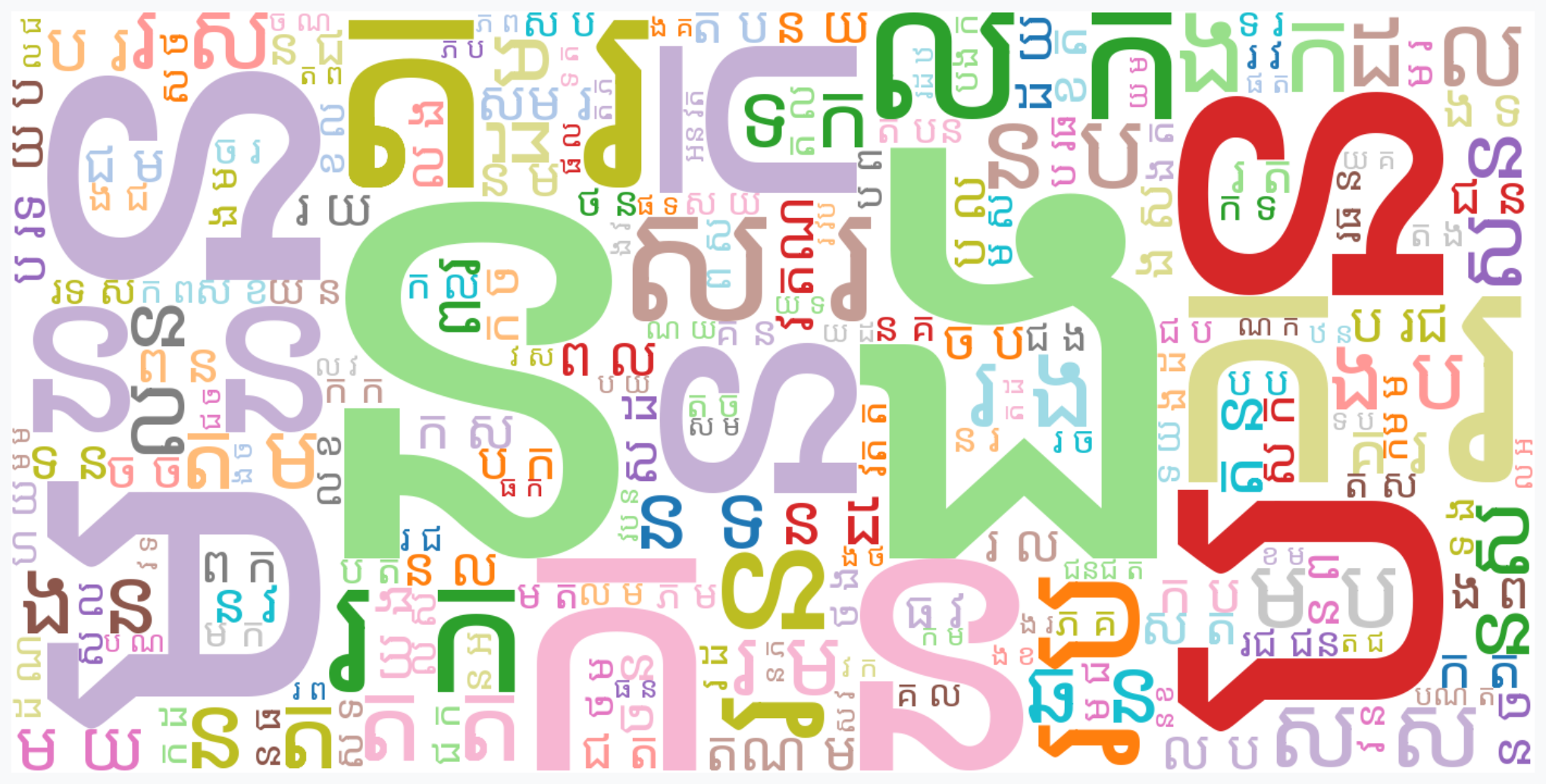}
    \caption{Word cloud of most frequent tokens in the Khmer corpus, written in Khmer script.}
    \label{fig:wordcloud_khmer}
\end{figure}
\begin{figure}[ht]
    \centering
    \includegraphics[width=\linewidth]{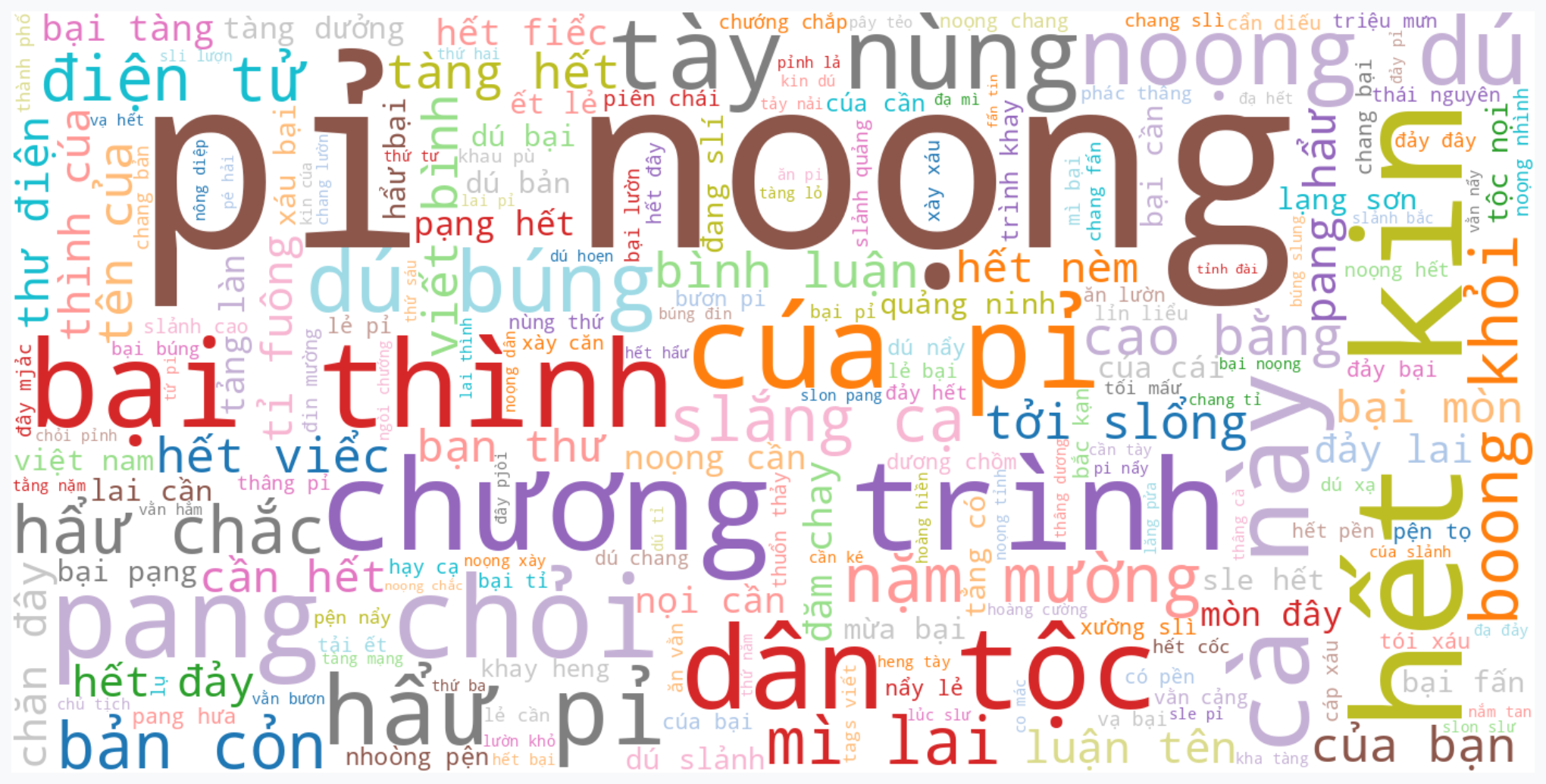}
    \caption{Word cloud of most frequent tokens in the Tay--Nung corpus, dominated by words such as \textit{noọng}, \textit{pỉ}, \textit{chương trình}, and \textit{dân tộc}.}
    \label{fig:wordcloud_tay_nung}
\end{figure}

\subsection{Domain Distribution in the Corpus}
\label{sec:domain-distribution}

In this section, we present the comprehensive distribution of the source domains found within our dataset across both the training and development sets. 
As shown in Table~\ref{tab:domain-distribution}, the corpus distribution is heavily dominated by national broadcasting platforms. Specifically, vovworld.vn accounts for the largest share (44.55\%), which is expected since it serves as the official external service targeted at global audiences. Together with VOV4 (vov4.vov.vn, 37.32\%), a dedicated channel for ethnic minority languages, these two sources contribute over 81\% of the entire dataset. 

In contrast, regional and provincial outlets represent a significantly smaller portion. Local sources like \textit{Can Tho Newspaper} account for 14.90\%, while provincial portals such as \textit{Ca Mau Newspaper} (1.21\%) and \textit{An Giang Newspaper} (0.57\%) face extreme data scarcity. This distribution highlights that centralized national media remains the primary resource for digital ethnic minority language data.

\begin{table}[ht]
\centering
\small 
\setlength{\tabcolsep}{4pt} 
\begin{tabular}{lrr}
\toprule
\textbf{Domain} & \textbf{\#URLs} & \textbf{\%} \\ 
\midrule
vovworld.vn & 19,765 & 44.55\% \\
vov4.vov.vn & 16,559 & 37.32\% \\
baocantho.com.vn & 6,609 & 14.90\% \\
km-dantocmiennui.baotintuc.vn & 646 & 1.46\% \\
khmer.baocamau.vn & 537 & 1.21\% \\
angiang.gov.vn & 251 & 0.57\% \\ 
\midrule
\textbf{Total} & \textbf{44,367} & \textbf{100.00\%} \\ 
\bottomrule
\end{tabular}
\caption{The domain distribution found in the corpus.}
\label{tab:domain-distribution}
\end{table}

\section{The influence of vocabulary on Rembert's performance}

\label{app:vocab-influence}

The results in Table~\ref{tab:classification_cktn} report all baselines --
including RemBERT -- \textit{after} vocabulary augmentation (Section
3.2), so the gap between RemBERT and CKTN-ELECTRA there isolates the
effect of the calibrated RTD objective (Section~\ref{sec:calibrated-rtd})
alone, holding vocabulary fixed. To further isolate the contribution of
vocabulary augmentation itself, we run an additional ablation: continued
MLM pretraining of the \textit{original, non-augmented} RemBERT tokenizer
directly on the CKTN corpus, under the same optimization budget and
downstream fine-tuning protocol as Section~\ref{sec:experimental-setup},
with no calibrated RTD involved.

\begin{table}[ht]
\centering
\small
\setlength{\tabcolsep}{4pt}
\begin{tabular}{l|cc}
\toprule
\textbf{Metric} & \textbf{No Vocab Aug.} & \textbf{Vocab Aug.\ (Table~\ref{tab:classification_cktn})} \\
\midrule
\multicolumn{3}{c}{\textit{Category Classification}}\\
\midrule
Accuracy $\uparrow$   & 0.0304 & 0.1454 \\
Macro-F1 $\uparrow$   & 0.0012 & 0.0053 \\
\midrule
\multicolumn{3}{c}{\textit{Information Retrieval}}\\
\midrule
MRR@10 $\uparrow$     & 0.7857 & 0.7787 \\
Recall@10 $\uparrow$  & 0.9090 & 0.9049 \\
\bottomrule
\end{tabular}
\caption{Effect of vocabulary augmentation on RemBERT after CPT, with no
calibrated RTD in either setting. ``Vocab Aug.'' reproduces the RemBERT row
from Table~\ref{tab:classification_cktn} (which already uses the extended
tokenizer of Section 3.2) for direct comparison.}
\label{tab:rembert-vocab-ablation}
\end{table}

\paragraph{Vocabulary augmentation helps, but only marginally.}
Extending RemBERT's vocabulary before CPT roughly doubles classification
Accuracy (0.0304~$\rightarrow$~0.1454) and more than quadruples Macro-F1
(0.0012~$\rightarrow$~0.0053). This confirms that lexical fragmentation is
part of RemBERT's failure mode: without an extended vocabulary, CKTN text
is split into long continuation-piece sequences (Table~\ref{tab:tokenizer}
shows RemBERT's un-augmented continuation-token ratio reaching 44.5\% on
Tay-Nung), leaving even less standalone lexical signal for the 256-d
embedding table to exploit. However, both settings remain far below every
other baseline in Table~\ref{tab:classification_cktn} (XLM-R: 0.8169/0.5617;
mBERT: 0.8024/0.5054) and orders of magnitude below CKTN-ELECTRA
(0.9214/0.7103), which shares RemBERT's backbone and augmented vocabulary
but adds the calibrated RTD objective. Vocabulary augmentation alone is
therefore \textit{necessary but far from sufficient} to make RemBERT a
competitive encoder on CKTN; the RTD objective and its script-aware
calibration (Section~\ref{sec:calibrated-rtd}) account for the
overwhelming majority of CKTN-ELECTRA's gain over plain RemBERT CPT, not
the vocabulary extension by itself.

\paragraph{Retrieval is insensitive -- and mildly \textit{anti}-correlated
-- with vocabulary quality.}
In contrast to classification, retrieval performance is essentially
unaffected by vocabulary augmentation, and if anything is marginally
\textit{higher} without it (MRR@10: 0.7857 vs.\ 0.7787; Recall@10: 0.9090
vs.\ 0.9049). This is consistent with the pattern already observed in
Section~\ref{sec:extrinsic} and the ablation study: because our retrieval
task pairs each summary with its own source article, it can largely be
solved through shared surface forms between query and document, regardless
of how cleanly the tokenizer segments those forms into semantic units. A
model does not need well-formed, low-fragmentation subwords to match
overlapping character sequences; it only needs the \textit{same}
fragments to appear on both sides of the pair, which happens whether or
not the vocabulary was extended. This result reinforces our central
methodological claim: lexical-overlap retrieval and language-modeling loss
are unreliable proxies for representation quality in script-heterogeneous,
low-resource adaptation, whereas category classification -- which requires
generalizing across lexically dissimilar documents within a class --
exposes real differences in tokenization and pretraining quality that
retrieval conceals.

\section{Sociolinguistic Context of Minority Languages in Vietnam}
\label{app:sociolinguistics}

\begin{figure}[ht]
    \centering
    \includegraphics[width=\linewidth]{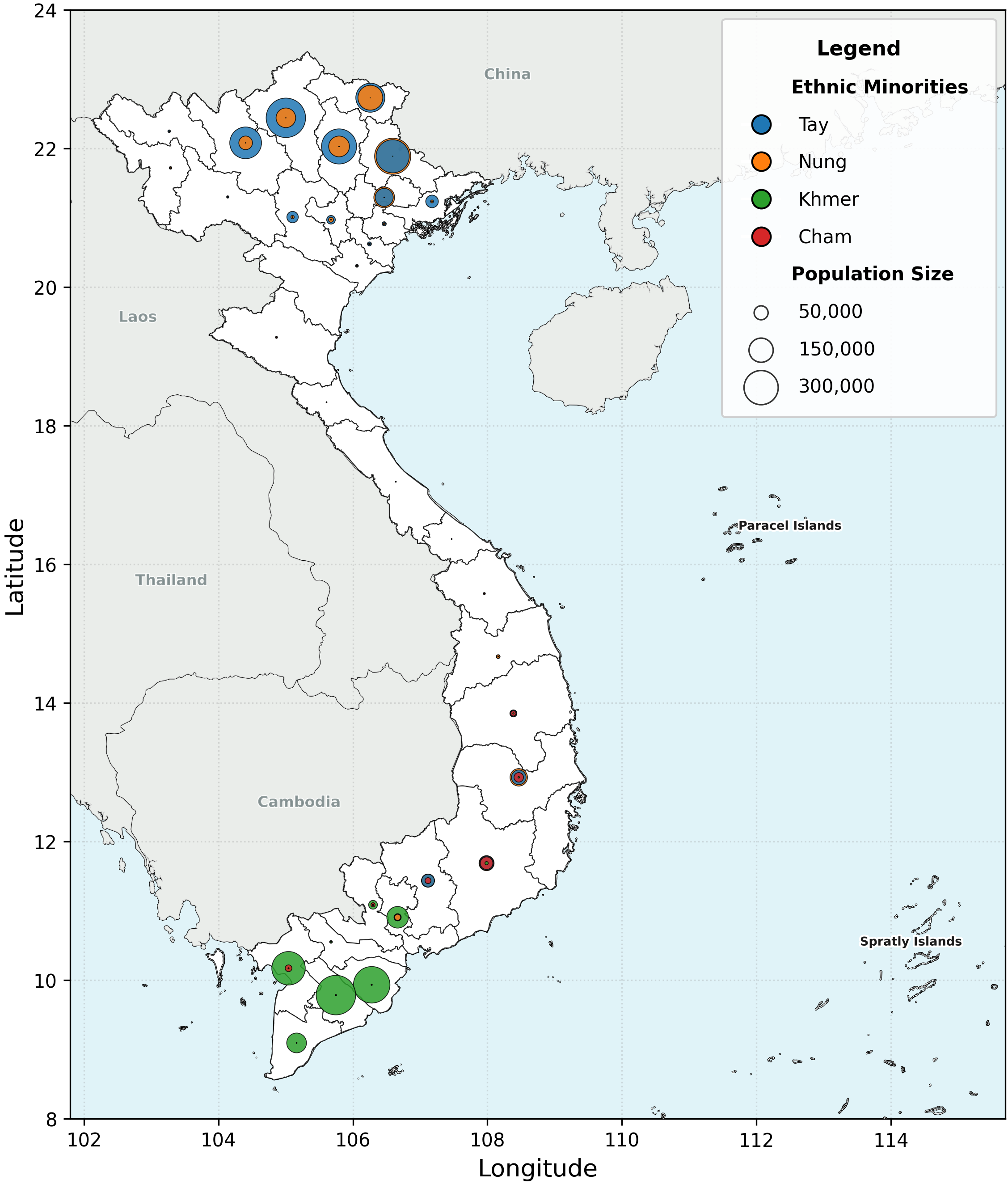}
    \caption{Geographic distribution and population sizes of selected ethnic minority communities (Tay, Nung, Khmer, and Cham) in Vietnam, covering the geographic region from latitudes 8.0°N to 24.0°N and longitudes 102.0°E to 115.5°E.}
    \label{fig:ethnic_map}
\end{figure}

\urldef\cemakhmerurl\url{http://www.cema.gov.vn/gioi-thieu/cong-dong-54-dan-toc/nguoi-khmer.htm}
\urldef\cemanungurl\url{http://www.cema.gov.vn/gioi-thieu/cong-dong-54-dan-toc/nguoi-nung.htm}
\urldef\cematayurl\url{http://www.cema.gov.vn/gioi-thieu/cong-dong-54-dan-toc/nguoi-tay.htm}
\urldef\cemachamurl\url{http://www.cema.gov.vn/gioi-thieu/cong-dong-54-dan-toc/nguoi-cham.htm}

From a taxonomic and geographical perspective, the ethnolinguistic landscape of the Khmer, Cham, and Tay-Nung communities in Vietnam is highly diverse. The Khmer in southern Vietnam, whose dialects are classified within the Central Khmer construct of the Austroasiatic family, predominantly inhabit the Mekong Delta region in southwestern Vietnam, clustering across inland plains, saline coastal areas, and southwestern border hills\footnote{\label{fn:cema-khmer}\cemakhmerurl}. The Cham, speaking an Austronesian language of the Western-Malayo-Polynesian branch, reside primarily in the south-central coastal provinces of Ninh Thuan\footnote{Effective July 1, 2025, under Resolution No. 202/2025/QH15, Ninh Thuan Province is merged with Khanh Hoa Province into the newly consolidated Khanh Hoa Province.} and Binh Thuan\footnote{Effective July 1, 2025, under Resolution No. 202/2025/QH15, Binh Thuan Province is merged with Dak Nong and Lam Dong provinces into the newly consolidated Lam Đong Province.}, traditionally occupying ground-level dwellings\footnote{\label{fn:cema-cham}\cemachamurl}. Conversely, the Tay and Nung languages are classified within the Kra-Dai (Tai-Kadai) language family~\cite{NST}. These communities are geographically distributed across the northeastern mountainous valleys, spanning provinces from Quang Ninh, Bac Giang\footnote{Effective July 1, 2025, under Resolution No. 202/2025/QH15, Bac Giang Province is merged with Bac Ninh Province into the newly consolidated Bac Ninh Province.}, and Lang Son to Lao Cai and Yen Bai\footnote{Effective July 1, 2025, under Resolution No. 202/2025/QH15, the entirety of Yen Bai Province and Lao Cai Province are merged into the newly consolidated Lao Cai Province.}, where the Tay typically reside in highly populated villages\footnote{\label{fn:cema-tay}\cematayurl}, while the Nung frequently live in interspersed settlements alongside them\footnote{\label{fn:cema-nung}\cemanungurl}.

From a macro-sociolinguistic perspective, the vitality of Khmer in Vietnam is increasingly restricted to specific localized domains, primarily sustained through the intergenerational transmission within family compounds and ritualized engagement in Theravada Buddhist religious practices~\cite{Taylor2014}. In formal administrative and educational settings, the dominant Vietnamese language acts as the primary medium of instruction and official communication, facilitating a high frequency of code-mixing among the Khmer community. However, rather than leading to language attrition, this contact has catalyzed a stable and directed bilingualism, where the younger generation, particularly those born after 1975, demonstrates enhanced proficiency in both Khmer and Vietnamese compared to their elders, reflecting a trajectory of sustainable linguistic development. Although Southern Khmer in Vietnam is generally classified within the Central Khmer dialect continuum and maintains mutual intelligibility with the source language, this variety exhibits highly significant lexical and phonological divergences~\cite{Kirby2021}. These distinctions largely stem from extensive language contact with Vietnamese, which has driven profound shift trajectories such as monosyllabization. Notably, in stark contrast to the fundamentally non-tonal nature of Standard Khmer, certain Khmer dialects in Vietnam (exemplified by the Kien Giang\footnote{Effective July 1, 2025, under Resolution No. 202/2025/QH15, Kien Giang Province is merged with An Giang Province into the newly consolidated An Giang Province.} variety) have developed a nascent pitch-based contrast system under the long-term influence of a dominant tonal language~\cite{Kirby2017}. The prominent phonological and lexical drift of these minority varieties from their genealogical roots inherently challenges traditional comparative assumptions and creates substantial barriers for standardizing text corpora.

The Eastern Cham communities exhibit a complex sociolinguistic framework defined by pervasive bilingualism with Vietnamese and an internal diglossia. This diglossia manifests through the coexistence of a conservative High (H) formal variety and a colloquial Low (L) variety, which has undergone severe structural reduction and lost almost all presyllables to become virtually monosyllabic. Furthermore, extensive contact with the dominant Vietnamese culture has led to significant lexical replacement and syntactic restructuring within the spoken language. Beyond sociolinguistic structures, the phonetic architecture of Eastern Cham is characterized by a phonological register system that originally emerged from the neutralization of voicing in onset stops~\cite{Brunelle2009}. Crucially, Eastern Cham demonstrates a hyper-salient distinction between its registers, which is directly attributed to its speakers' quasi-native bilingualism in Vietnamese. This phenomenon suggests a fine-grained phonetic convergence where existing indigenous register properties are significantly altered and reshaped by contact with a dominant tonal language.

The sociolinguistic ecology and the distribution of writing systems among the Cham community reveal a complex interplay of religious ideology, educational policy, and technological constraints. It is an oversimplification to characterize the Eastern Cham as exclusively Hindu, as part of the population practices a syncretic form of Islam and is known as the Pani~\cite{Linh2022ChamAhier}. While the traditional Indic script, akhăr thrah, is preserved for formal religious ceremonies and the transcription of classical literature, the Pani subgroup concurrently utilizes an Arabic-based script, akhăr pani, specifically for their internal religious rituals. Furthermore, while primary education programs in Cham-speaking areas are strictly dedicated to the instruction and preservation of the traditional akhăr thrah script~\cite{Sang2015ChamFont}, practical modern communications such as television broadcasts and digital messaging predominantly rely on an ad hoc Romanized script due to immediate technological limitations. Consequently, the community maintains a staunch opposition to any official Romanization efforts, viewing such initiatives as a direct threat to the preservation of their cultural heritage.

In contrast to the severe sociolinguistic marginalization experienced by many micro-minority groups, the Tay and Nung languages represent a distinct dynamic within the ethnolinguistic landscape of northern Vietnam. Boasting a relatively large demographic base (with the Tay population reaching over 1.8 million \footref{fn:cema-tay} and the Nung exceeding 1 million\footref{fn:cema-nung}), both languages benefit from officially recognized writing systems and have been supported by foundational linguistic documentation, such as formal grammar texts and multilingual dictionaries published in the 1970s and 1980s. The functional vitality of Tay is further reinforced by its active utilization in national mass media, including designated state television and radio broadcasts. Crucially, from a sociolinguistic perspective, Tay and Nung operate with a relatively high status compared to other indigenous languages in the region. Their regional prestige exerts direct assimilation pressure on highly endangered communities, a phenomenon exemplified by the language shift observed in the Red Co Lao group, whose members have transitioned to speaking Tay alongside Chinese.

However, this robust demographic base does not automatically guarantee full functional security within formal domains. In the broader context of Vietnam's national education policy, the state system overwhelmingly prioritizes Vietnamese as the primary language of instruction, effectively relegating minority languages to a limited, supplementary, or extracurricular status~\cite{languageEducationPolicy}. Sociolinguistic evaluations indicate that such systemic limitations heavily restrict the development of formal written literacy among ethnic minority youth, including the Tay-Nung communities. As a result, the lack of extensive utilization in formal administrative and standardized educational frameworks channels these languages primarily into oral, informal, and domestic domains, thereby constraining their capacity to establish a strong functional presence in non-oral public spheres.

Accordingly, formal written communication remains a challenge for most minority languages, which are largely confined to oral, informal, and rural in-group domains~\cite{Thong2019}. As a sub-set of these minority tongues, Tay and Nung mirror this broader trend; the resulting lack of active, standardized written production in official spheres fundamentally obstructs the development of large-scale text corpora, thereby firmly entrenching them as acutely low-resource languages in terms of documented literacy. The drastic phonological and lexical drift of these minority varieties from their genealogical counterparts, such as the independent tonal developments in Kien Giang Khmer~\cite{Kirby2017} and the severe structural reduction in Eastern Cham, challenges traditional comparative assumptions. Furthermore, linguistic documentation tasks are severely impeded by extreme data sparsity and orthographic instability \cite{DBLP:journals/corr/abs-2106-15115}. The pervasive code-mixing observed among younger Khmer generations~\cite{Dinh2015}, coupled with the fragmented, ad hoc Romanization of Cham in digital messaging due to localized hardware and font constraints~\cite{Brunelle2008}, inherently obstructs the construction of standardized, large-scale textual repositories. Consequently, this entire linguistic cluster suffers from a critical deficit of standardized written data, firmly entrenching them as digitally vulnerable and under-documented languages \cite{joshi-etal-2020-state}.

\section{Reproducibility and Implementation Details}
\label{app:reproducibility}

This appendix documents the software environment, preprocessing,
training objectives, optimization settings, and evaluation procedures
used in our experiments.

\paragraph{Software environment.}
All models are implemented in PyTorch~2.12.1 using
Transformers~5.12.1 and NumPy~2.4.6. Tokenizer augmentation uses
SentencePiece~0.2.1. Category-classification metrics are computed with
scikit-learn~1.9.0. The current implementation runs in full precision
and does not use mixed-precision training.

\paragraph{Tokenization and input processing.}
We use the slow \texttt{RemBertTokenizer}, which is backed by
SentencePiece. After vocabulary augmentation, the tokenizer contains
254,513 entries. All pretraining and downstream scripts truncate or
pad inputs to a maximum sequence length of 512 tokens.

The current repository does not apply an external Khmer word segmenter.
In particular, neither \texttt{khmer-nltk} nor \texttt{khmernltk} is
installed or invoked by the released scripts. Khmer documents are read
directly from the normalized JSON files and tokenized by the
SentencePiece-backed model tokenizer. We state this explicitly to
distinguish the released implementation from preprocessing variants
that use separate Khmer word segmentation.

\paragraph{Continued pretraining.}
The MLM baselines and CKTN-ELECTRA are trained with AdamW using a
learning rate of $2\times10^{-5}$, weight decay of $0.01$, a warmup
ratio of $0.06$, and gradient clipping with maximum norm $1.0$. The
learning rate follows a linear warmup and decay schedule. The masking
rate is $0.15$, and the maximum input length is 512.

CKTN-ELECTRA jointly optimizes the generator MLM objective and the
discriminator RTD objective. The discriminator-loss coefficient is set
to zero during epochs 1--2, increased linearly to
$\lambda_{\max}=50$ by epoch 3, and held fixed thereafter. Replacement
candidates are selected from the generator's top 64 predictions using
temperature $\tau=1.25$. Candidates are filtered by script and surface
form, and their cosine similarity with the original token must fall
within $[0.15,0.95]$.

\paragraph{Category classification.}
Category classifiers are fine-tuned for five epochs with batch size 32,
learning rate $10^{-4}$, and warmup ratio $0.06$. The implementation
supports either the final-layer \texttt{[CLS]} representation or masked
mean pooling over non-padding tokens. Weighted cross-entropy is enabled
by default to account for label imbalance. We report accuracy and
Macro-F1 using scikit-learn, with
\texttt{zero\_division=0}. Micro-F1 is also computed for diagnostic
purposes.

\paragraph{Information retrieval.}
Retrieval is implemented as a bi-encoder in PyTorch. Queries and
documents are encoded independently using masked mean pooling, followed
by $\ell_2$ normalization. Similarity is computed by a dot product,
which is equivalent to cosine similarity after normalization. The
current implementation performs exact similarity computation and does
not construct a FAISS index.

The query is the article summary when available, with source tags used
only as a fallback for missing summaries. The document input is formed
by concatenating the title and article content. Retrieval fine-tuning
uses AdamW for five epochs with batch size 16, evaluation batch size 16,
learning rate $2\times10^{-5}$, and contrastive temperature $0.1$.
Batches are grouped by source to reduce the risk that trivial
source-specific cues dominate negative sampling. We report MRR@10 and Recall@10.
\end{document}